\documentclass{article}

     \PassOptionsToPackage{numbers, compress}{natbib}
\usepackage[preprint]{neurips_2021_imagenet}
\usepackage{multirow}

\usepackage[utf8]{inputenc} % allow utf-8 input
\usepackage[T1]{fontenc}    % use 8-bit T1 fonts
\usepackage{hyperref}       % hyperlinks
\usepackage{url}            % simple URL typesetting
\usepackage{booktabs}       % professional-quality tables
\usepackage{amsfonts}       % blackboard math symbols
\usepackage{nicefrac}       % compact symbols for 1/2, etc.
\usepackage{microtype}      % microtypography
\usepackage{xcolor}         % colors
\usepackage{authblk}

\title{One Pass ImageNet}

\author[1]{\bf Huiyi Hu}
\author[1]{\bf Ang Li}
\author[2]{\bf Daniele Calandriello}
\author[1]{\bf Dilan Gorur}

\affil[1]{DeepMind, Mountain View, USA}
\affil[2]{DeepMind, Paris, France}

\affil[ ]{\vskip -1em\texttt {\{clarahu,anglili,dcalandriello,dilang\}@deepmind.com}}

\begin{document}

\maketitle

\begin{abstract}
  We present the One Pass ImageNet (OPIN) problem, which aims to study the effectiveness of deep learning in a streaming setting. 
  ImageNet is a widely known benchmark dataset that has helped drive and evaluate recent advancements in deep learning. Typically, deep learning methods are trained on static data that the models have random access to, using multiple passes over the dataset with a random shuffle at each epoch of training. Such data access assumption does not hold in many real-world scenarios where massive data is collected from a stream and storing and accessing all the data becomes impractical due to storage costs and privacy concerns. For OPIN, we treat the ImageNet data as arriving sequentially, and there is limited memory budget to store a small subset of the data. We observe that training a deep network in a single pass with the same training settings used for multi-epoch training results in a huge drop in prediction accuracy. We show that the performance gap can be significantly decreased by paying a small memory cost and utilizing techniques developed for continual learning, despite the fact that OPIN differs from typical continual problem settings. We propose using OPIN to study resource-efficient deep learning.
    % below is too much detail. 
  %We observe that training  in one pass a Residual network with 50 layers results in only $31.5\%$ top-1 accuracy, a significant drop from the $76.9\%$ accuracy obtained from the commonly adopted 90-epoch training. Inspired by recent advances in continual learning, we study a solution that utilizes an Error-Prioritized Replay (EPR) memory and trains the model with additional replay examples at each step. The proposed approach achieves $62.2\%$ top-1 accuracy in the validation set while reducing total back-propagation computation by 93.3\% and required data storage by 90\%. While there is still performance gap between one pass training and multi-epoch training, the results show a promising research direction on resource-efficient deep learning from large scale real world data.
\end{abstract}

\section{Introduction}

ImageNet \cite{imagenet_cvpr09} is one of the most influential benchmarks that has helped the progress of machine learning research.  Tremendous progresses have been made over the past decade in terms of a model's accuracy on the ImageNet dataset. While improving the ImageNet accuracy has been the major focus of the past, little effort has been made on studying the resource efficiency in ImageNet supervised learning. Most existing supervised learning methods assume the data is i.i.d., static and pre-existing. They train models with multiple epochs, i.e., multiple passes of the whole dataset. Specifically, a top-performing Residual Neural Network \cite{He2016DeepRL} is trained with 90 epochs; the model needs to review each of the 1.2M examples 90 times. One natural question to ask is whether it is necessary to train a model with so many passes over the whole data.

We are also motivated by the fact that real world data often comes hourly or daily in a stream and in a much larger scale. Maintaining all the data in a storage can be expensive and probably unnecessary. Additionally, real world data sometimes contains private information from human users, which further restricts the possibility of saving all the data into a separate storage. Without a pre-recorded dataset, the popular multi-epoch training method becomes impractical to these real world scenarios. 

We propose the One Pass ImageNet (OPIN) problem to study the resource efficiency of deep learning from a streaming setting with constrained data storage, where space complexity is considered an important evaluation metric.
%The problem is built upon ImageNet, whose data is treated as coming from a stream. 
The goal is to develop a system that trains a model where each example is passed to the system only once. 
There is a small memory budget but no restriction on how the system utilizes its memory; it could store and revisit some examples but not all. Unlike the task-incremental continual learning setting \cite{masana2021classincremental, taskonomycl}, the One-Pass ImageNet problem does not have a special ordering of the data, nor is there a specific distribution shift as in task-free continual learning \cite{cai2021online,aljundi2019taskfree,DBLP:journals/corr/abs-2106-12772}. That is, the data comes from a fixed uniform random order. We use ResNet-50 \cite{He2016DeepRL} in all our experiments, and leave the question of choice of architecture as future work.

%We note that a simple 90-epoch training of a ResNet-50 \cite{He2016DeepRL} leads to 76.9\% top-1 accuracy.
We observe that training a ResNet-50 \cite{He2016DeepRL} in a single pass leads to only 30.6\% top-1 accuracy on the validation set, a significant drop from 76.9\% top-1 accuracy obtained from the common 90-epoch training\footnote{The top-1 accuracy is obtained with 1-crop evaluation in ImageNet.}. Inspired by the effectiveness of memory-based continual learning \cite{taskonomycl,buzzega2020rethinking,rolnick2019experience,bangKYHC21}, we propose an error-prioritized replay (EPR) method to One Pass ImageNet. The proposed approach utilizes a priority function based on predictive error. Results show that EPR achieves 65.0\% top-1 accuracy, improving over naive one-pass training by 34.4\%. Although it still performs 11.9\% lower than multi-epoch training in terms of accuracy, EPR shows superior resource efficiency which reduces total gradient update steps by 90\% and total required data storage by 90\%. 

We believe OPIN is an important first step that allows us to understand how existing techniques can train models in terms of computation and storage efficiency, although it may not be the most realistic example for the data streaming setting. We hope our results could inspire future research on large scale benchmarks and novel algorithms on resource-efficient supervised learning.

\section{One Pass ImageNet}
The One Pass ImageNet problem assumes that examples are sent in mini-batches and do not repeat. The training procedure ends when the whole dataset is revealed. No restriction is applied on how the trainer utilizes its own memory, so a memory buffer that records past examples is allowed. However, the amount of data storage is a major evaluation metric considered as space efficiency.

We perform our study on a commonly used ImageNet solution: A ResNet-50 \cite{He2016DeepRL} trained over 90 epochs with cosine learning rate and augmented examples. We refer to this method as \textit{Multi-epoch} throughout the paper. The images are preprocessed by resizing to $250\times250$ and then performing augmentation into a size of $224\times224$. During training, the augmentation includes random horizontal flipping and random cropping. At test time, only center cropping is applied to the images.

\begin{table}
\setlength{\tabcolsep}{9pt}
\caption{Comparing One-Pass methods with Multi-epoch training in accuracy, storage, compute metrics. Naive One-Pass method results in significant accuracy drop. A priority replay based method achieves better accuracy while maintaining the significant improvements on storage and compute metrics over multi-epoch training. Higher accuracy, lower storage and lower compute are better.}
\label{tab:main}
\centering
\begin{tabular}{ cccccc  }
 \toprule
  & \bf Accuracy (\%) $\uparrow$ & \bf Storage (\%) $\downarrow$  & \bf Compute (\%) $\downarrow$\\
 \midrule
\bf Multi-epoch (90 epochs) & 76.9 & 100 & 100\\
\bf One-Pass (Naive)   & 30.6 & 0 & 1.1 \\
\bf One-Pass (Prioritized Replay) & 65.0 & 10 & 10 \\
 \bottomrule
\end{tabular}
\end{table}

\noindent\textbf{Evaluation metrics.}
While standard ImageNet benchmark focuses on a model's overall accuracy, the One-Pass ImageNet problem aims at studying the learning capability under constrained space and computation. So the problem becomes essentially a multi-objective problem. We propose to evaluate training methods using three major metrics: (1) \textit{accuracy}, represented by the top-1 accuracy in the test set, (2) \textit{space}, represented by total additional data storage needed, and (3) \textit{compute}, represented by the total number of global steps for back-propagation. The space and compute metric is calculated relative to the multi-epoch training method, i.e., both metrics for Multi-epoch method are 100\%. The Multi-epoch method needs to save all the data into storage, so the space metric is measured by the size of the data storage divided by the size of the dataset. The Multi-epoch method needs to train a model with 90 epochs (or 100M global steps), so the compute metric is measured by the total number of back-propagation operations divided by 100M.

\noindent\textbf{Naive baseline.} A simple baseline method for the One Pass problem is to train a model with the same training configuration that multi-epoch training uses but with only a single epoch, which we call \textit{Naive One-Pass}. Since each example is seen  only once, we replace the random augmentation with center cropping (which is used in the model evaluation) in this Naive baseline. Table \ref{tab:main} shows a comparison between multi-epoch training and naive one-pass training measured in three metrics: accuracy, space and compute. All metrics are in percentage. While Naive One-Pass is significantly worse than multi-epoch in terms of accuracy, its space and compute efficiency are both significantly higher. The Naive One-Pass does not need to save any data examples into memory and at the same time it only trains for one-epoch, the total number of training steps is $1/90\approx 1.1\%$ that of multi-epoch training.

\noindent\textbf{Problem Characteristics.} Here we list four properties of the OPIN problem as below:
\begin{enumerate}
    \item \textit{The cold-start problem}: Model start from random initialization. So the representation learning becomes challenging in OPIN especially during the early stage of the training. 
    \item \textit{The forgetting problem}: Each example is passed to the model only once. Even though the data is i.i.d., vanilla supervised learning is likely to incur forgetting of early examples.
    \item \textit{A natural ordering of data}: No artificial order of the data is enforced. So the data can be seen as i.i.d., which is different from many existing continual learning benchmarks.
    \item \textit{Multiple objectives}: The methods are evaluated using three metrics (accuracy, space and compute), so the goal is to improve all three metrics in a single training method.
\end{enumerate}

\section{A Prioritized Replay Baseline}
Memory replay is a common approach in continual learning \cite{rolnick2019experience}. Existing works have shown that memory replay is effective in sequential learning \cite{Hsu18_EvalCL, taskonomycl}. As our first investigation to the One Pass ImageNet problem, we study how a replay buffer could improve the overall performance in OPIN.

\subsection{Replay buffer}

Replay buffer is an extra memory that explicitly saves the data. This memory usually has a very limited size. At each training step, the received mini-batch of examples is inserted into this replay memory. Since the buffer size is smaller than the whole dataset, the typical solution is to apply the reservoir sampling strategy \cite{reservoirsampling} where each example is inserted at a probability of $p(n,m)=m/n$, where $m$ is the memory size and $n$ the total number of seen examples. In order to introduce a favor on fresh examples\footnote{Similar ideas on encouraging recent examples in reservoir sampling can be found in \cite{biased-reservoir,osborne-etal-2014-exponential}.}, we incorporate a factor to the inclusion probability, i.e., $p(n,m)=\beta m/n$ (we choose $\beta=1.5$), so that more recent examples will be included in the memory.

At each training step, extra examples are sampled from the replay buffer. These examples are trained by the model together with the incoming mini-batch. Existing research on continual learning has shown that uniform sampling from a replay buffer is effective in many cases \cite{taskonomycl}, however, a common intuition is that examples are not equally important when being replayed. The idea of prioritized experience replay \cite{schaul2016prioritized} is to add a priority score to each example and sample from the buffer according to the probability distribution normalized from the priority scores.

In order to apply  data augmentation  to replay examples, we save jpeg-encoded image bytes into the replay buffer instead of image tensors, which turns out to be more space efficient (3x more examples can be saved under the same memory budget).
We found replaying multiple examples at each step could dramatically improve the model accuracy, although with a trade-off in compute. For each step when we receive one mini-batch of images, we replay $k$ mini-batches from the replay buffer, which leads to $k+1$ epochs of compute effectively.

\subsection{Priority function}
We study a prioritized replay method that uses the predictive error (loss value) as the priority function, which we call Error-Prioritized Replay (EPR). The priority function is defined as 
\begin{equation}
    P(x,y)=1-\alpha e^{-\ell(x,y;\theta)}
\end{equation}
where $\ell(x,y;\theta)$ is the loss value given input example $x$, ground truth label $y$ and model parameter $\theta$.  
The smoothing factor $\alpha$ varies from 0 to 1 such that $\alpha(T)=1-cos(T/T_\textrm{max})$ where $T$ is the current global step and $T_\textrm{max}$ is the maximum global step. We choose a smoothed $\alpha$ because the model's prediction is not trustworthy at the early stage of training. When the loss function is cross-entropy $\ell(x,y;\theta)=-\log f_y(x;\theta)$, it can also be shown that the priority value $P(x,y)=1-\alpha f_y(x;\theta)$, which is made of the model's confidence on the ground truth label.

\subsection{Importance weight}
The examples sampled from the priority replay buffer changes the data distribution. To simplify the notation, we omit the label $y$ from the equations in this section. Let $p(x)$ be the distribution of the original data and $q(x)$ be the distribution in the replay buffer. The original objective is $\mathbb E_p[\ell(x;\theta)]$. Supposing $k$ mini-batches are sampled from the replay buffer at each step, directly combining replay examples and current examples will lead to a minimization of $\mathbb E_p[\ell(x;\theta)]+k\mathbb  E_q[\ell(x;\theta)]$. In order to correct the distribution shift, we use an importance weight $w(x)=p(x)/q(x)$ to each replay example because
$\mathbb E_q[w(x)\ell(x;\theta)]=\mathbb E_p[\ell(x;\theta)]$.
Given that the original distribution can be assumed uniform ($p(x)=1$),  the importance weight of each replay example is $w(x)\propto 1/P(x)$, inversely proportional to its priority value. The weights of each mini-batch are normalized to mean 1. 

%$w(x) \propto 1/Priority(x)$

\section{Experiments}

\subsection{Experimental setup}
The model in our study is a Residual Neural Network \cite{He2016DeepRL} with 50 layers (\textit{aka.} ResNet-50). We use cosine learning rate decay for all experiments, with initial learning rate 0.1. The model is optimized using stochastic gradient descent with Nesterov momentum 0.9. The batch size is 128. We evaluate different approaches with 10 different random orders of the data.

%\begin{table}
%\setlength{\tabcolsep}{12pt}
%\caption{Priority functions at 3\% replay capacity}
%\label{tab:priority}
%\centering
%\begin{tabular}{ cccc  }
% \toprule
%  & \bf uniform & \bf uncertainty-based & \bf loss-based \\
%  & $p(x)=1$ & $p(x)=1-cf_\text{max}(x)$ & $p(x)=1-ce^{-\ell(x)}$\\
% \midrule
%\bf w/o priority update   & \\
%\bf w/ priority update & \\
% \bottomrule
%\end{tabular}
%\end{table}

\begin{table}
\setlength{\tabcolsep}{16pt}
\caption{One-Pass ImageNet evaluation: Computation, Memory and Accuracy. The effective epoch is computed as the number of replay steps added by 1. The compute metric is calculated by the effective epoch divided by 90 (compared against the 90-epoch training). The multi-epoch results on the last column is obtained by training with the same number of effective epochs with cosine learning rate decay (however, it requires saving the full dataset to storage).}
\label{tab:evaluation}
\centering
\begin{tabular}{ c|cccc|c  }
 \toprule
  Effective& \multirow{ 2}{*}{Computation}   & \multicolumn{3}{c|}{Storage (Prioritized Replay)} & Multi-epoch\\
 Epochs&& 1 \% &5 \% &10 \%  & 100\% Storage\\
 \midrule
 2&$2/90\approx2.2\%$   &  44.7   & 45.1 & 45.7 & 46.1 \\
 4&$4/90\approx4.4\%$&  55.5   & 57.1  & 57.2 & 59.0\\
 6&$6/90\approx6.7\%$ & 58.9  & 61.3  & 62.2  & 64.1\\
 9&$9/90=10\%$& 59.3&63.2&65.0&68.2\\
 \bottomrule
\end{tabular}
\end{table}

\subsection{Results}
The results are shown in Table \ref{tab:evaluation}. We performed 9 experiments with the replay steps being 1, 3, 5, 8 and the size of replay buffer being 1\%, 5\% and 10\% of the dataset. The effective epoch is computed as adding the number of replay steps $k$ by 1 because at receiving each new mini-batch, $k$ mini-batches of the same size are sampled from the replay buffer. So effectively, $k+1$ back-propagation operations are performed when the model sees a new mini-batch. In addition to the One-Pass solutions, we also show the results of multi-epoch training with the same number of effective epochs. The learning rate decay is adjusted accordingly to decay to minimum at the end of the corresponding epoch. We observe multiple trends from this table:
\begin{enumerate}
    \item Prioritized replay with 10\% memory size achieves performance very close to the multi-epoch training method under the same computational cost. And the multi-epoch method utilizes the full dataset which requires a large data storage. It is unknown whether multi-epoch gives a performance upper bound, we believe it is a strong target performance to reference.
    \item 1\% data storage gives a strong starting point for prioritized replay. Having a 1\% data storage (equivalent to 100 mini-batches) dramatically improves the naive One-Pass performance by 28.7\%. According to the table, the accuracy increase from 1\% to 10\% storage is 1.0\% for 2 epochs, 1.7\% for 4 epochs, 3.3\% for 6 epochs and 5.7\% for 9 epochs. 
    \item When the buffer size becomes bigger, the accuracy gains more when the number of replay steps is more. From 5\% size to 10\% size, the model accuracy increase by 0.6\% and 0.1\% respectively for replay step 1 and 3, while the accuracy increases by 0.9\% for replay step 5. The model accuracy saturates quickly if one only increases either storage size or the replay steps. Increasing both of them could potentially incur much bigger accuracy boost.
\end{enumerate}

\subsection{Discussion}

%\begin{table}
%\setlength{\tabcolsep}{12pt}
%\caption{Comparison among three priority functions with 10\% data storage and 5 replay steps. The accuracy mean and standard deviation of multiple runs are shown. Although uniform replay buffer is a competitive baseline method, it suffers from higher accuracy variance, which is critical to model reproducibility. Results are computed over 10 random data orders.}
%\label{tab:priority}
%\centering
%\begin{tabular}{ c|ccc  }
% \toprule
% \multirow{2}{*}{Accuracy} &uniform& static error-based & dynamic error-based \\
%  &$p(x,y)=1$& $p(x,y)=1-e^{-\ell(x,y)}$& $p(x,y)=1-\alpha e^{-\ell(x,y)}$\\
% \midrule
% Mean & 61.82 & 61.84 & 62.17\\
% Std & 0.685 & 0.574 & 0.286\\
% \bottomrule
%\end{tabular}
%\end{table}

\noindent\textbf{Priority function.} Like recent literature \cite{taskonomycl,chaudhry2020using} suggested, we also observe competitive results using vanilla replay (with a uniform sample memory) in the One-Pass ImageNet problem. Specifically, a uniform memory leads to 64.7\% top-1 accuracy with 8 replay steps and 10\% storage, slightly worse than the prioritized replay method (65.0\%). The standard errors of mean for both are 0.07\%. Although the improvement is small, we believe the potential of prioritized replay is not fully explored in this problem and we leave the study of a better designed prioritization as future research.

\noindent\textbf{Importance weights.} To study the effectiveness of importance weights applied to the loss function, we evaluate the prioritized replay without importance weight at 1\% storage. With 5 replay steps (6 effective epochs), no importance weight results in 58.0\% top-1 accuracy, 0.9\% lower than the one with importance weight. With 8 replay steps, no importance weight results in 58.3\% accuracy, 1.0\% lower than using importance weight. The reason of this accuracy gap is due to the fact that the distribution of examples in the replay buffer differs from that of the evaluation data set. 

\noindent\textbf{Priority updating.} An interesting and somewhat surprising result we obtained is that updating the priorities of the examples retrieved from the replay buffer does not lead to better accuracy. An experiment using 5 replay steps and 1\% storage shows that, after an example is replayed from the buffer, immediately updating its priority in the replay buffer using the latest model parameters results in a top-1 accuracy of 58.8\% while we can achieve 58.9\% without updating.

%Many recent literature has suggested vanilla experience replay being competitive in continual learning \cite{chaudhry2020using,taskonomycl}. We also observe the effectiveness of vanilla replay in the One-Pass ImageNet problem. It can be seen as using a uniform priority function under our framework. We compare three priority functions in Table \ref{tab:priority}: uniform, a static error-based function without the decay factor $\alpha$ and the proposed error-based function. We show the mean accuracy and the standard deviation of multiple runs. While we also find a uniform priority function produces close accuracy, its performance variance is much higher than the proposed dynamic error-based priority. The accuracy variance is critical since it measures the reproducibility of a training method. We also find that a static error-based priority function performs slightly worse than the proposed dynamic one. The major reason is because in the early stage of training, the model's prediction is unreliable which leads to noisy priority scores among different examples. The dynamic priority basically smoothes this effect by transitioning the priority scoring system from uniform in the early phase to error-based in the late phase of training.

%\subsection{The choice of importance weight}
%We compare with the importance weight used in prioritized DQN %\cite{schaul2015prioritized} where
%\begin{equation}
%    w(x)=\left(\frac{1}{Bp(x)}\right)^\beta,\quad \hat %w(x_i)=\frac{w(x_i)}{\max_{j=1}^B w(x_j)}
%\end{equation}
%and $\beta$ changes from 0.5 to 1 throughout the training.

\section{Related Work}
The OPIN problem is related to continual learning \cite{DBLP:journals/corr/abs-2106-12772,yin2020sola,borsos2020coresets,isele2018selective,van2020brain,chaudhry2018efficient,rusu2016progressive} and incremental learning \cite{Castro_2018_ECCV}. Existing continual learning methods are often evaluated in benchmarks composed of standard supervised learning datasets such as MNIST and CIFAR \cite{Hsu18_EvalCL,kirkpatrick2017overcoming,farajtabar2020orthogonal}. Recent efforts are exerted on benchmarks built upon large scale datasets such as ImageNet \cite{wu2019large} and Taskonomy \cite{taskonomycl}. Class-incremental ImageNet \cite{wu2019large} is probably the closest to our benchmark in continual learning which splits ImageNet into multiple tasks and each tasks consist of examples labeled as a certain subset of classes. The data for each task is given to the model altogether, so the model is able to repeat multiple passes for the data in the same task. OPIN differs in that it has no task concept and the data obeys a natural order instead of a manually designed class incremental ordering. 

The OPIN problem is also related to stream data learning \cite{journals/sigkdd/GomesRBBG19,Souza2020ChallengesIB,dieuleveut2016nonparametric} and online learning \cite{10.1109/TIT.2004.833339, 6842642,onlinesgd}. While there have been many benchmarks utilizing real world streaming data \cite{Souza2020ChallengesIB}, they often have smaller scale, less complex features and limited number of classes. Their solutions are often limited to linear or convex models. The One-Pass ImageNet problem could also contribute to the research of stream data learning from the perspective of deep neural architectures and large scale high dimensional data.

\section{Conclusion}
We presented the One-Pass ImageNet (OPIN) problem which aims at studying not only the predictive accuracy along with the space and compute efficiency of a supervised learning algorithm. The problem is motivated from real world applications that comes with a stream of extremely large scale data, which leads to impracticality of saving all the data into a dedicate storage and iterating the data for many epochs. We proposed an Error-Prioritized Replay baseline to this problem which achieves 65.0\% top-1 accuracy while reducing the required data storage by 90\% and reducing the total gradient update compute by 90\%, compared against the popular 90-epoch training procedure in ImageNet. We hope OPIN could inspire future research on improving resource efficiency in supervised learning.

\paragraph{Acknowledgement.} The authors would like to thank Razvan Pascanu, Murray Shanahan, Eren Sezener, Jianan Wang, Talfan Evans, Sam Smith, Soham De, Yazhe Li, Amal Rannen-Triki, Sven Gowal, Dong Yin, Arslan Chaudhry, Mehrdad Farajtabar, Timothy Nguyen, Nevena Lazic, Iman Mirzadeh, Timothy Mann and John Maggs for their meaningful discussions and contributions.

\bibliographystyle{unsrt}
\bibliography{references.bib}

\end{document}